\documentclass{article}
\usepackage{ijcai24}

\usepackage{times}
\usepackage{soul}
\usepackage{url}
\usepackage[hidelinks]{hyperref}
\usepackage[utf8]{inputenc}
\usepackage[small]{caption}
\usepackage{graphicx}
\usepackage{amsmath}
\usepackage{amsthm}
\usepackage{booktabs}
\usepackage{algorithm}
\usepackage{algorithmic}
\usepackage[switch]{lineno}

\usepackage{amssymb}
\usepackage{multirow}

\title{Learning Position-Aware Implicit Neural Network for Real-World Face Inpainting}

\author{
Bo Zhao\textsuperscript{\rm 1}
\and
    Huan Yang \textsuperscript{\rm 2}$^{*}$ \and
    Jianlong Fu\textsuperscript{\rm 2}
\affiliations
\textsuperscript{\rm 1} Beijing Institute of Technology,
\textsuperscript{\rm 2} Microsoft Research Asian\\
\emails
    1198327714@qq.com, 
    huayan@microsoft.com, 
    jianf@microsoft.com
}

\begin{document}

\twocolumn[{
\renewcommand\twocolumn[1][]{#1}%
\maketitle
\begin{center}
    \centering
    \includegraphics[height=5.7cm]{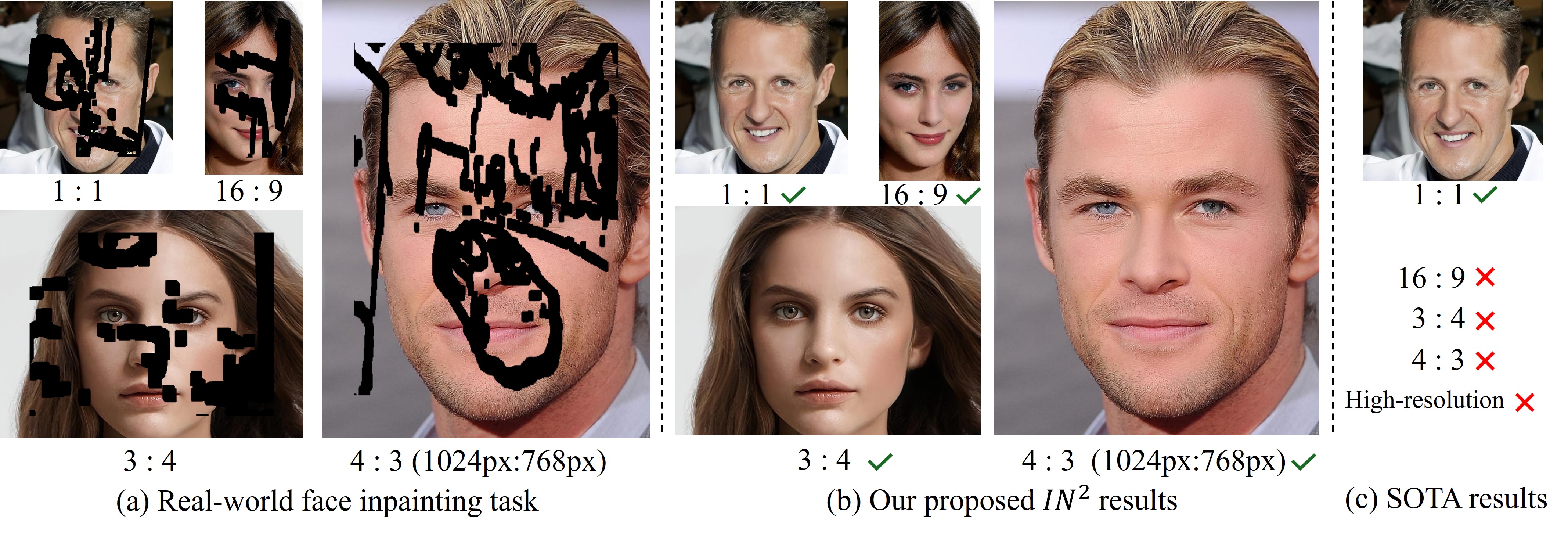}

    \captionof{figure}{\textbf{An Illustration of real-world face inpainting scenarios and comparison between our proposed methods and SOTA approaches.} (a) Examples of real-world inpainting images of different aspect ratios and resolutions. (b) Our proposed method can
process the input of arbitrary resolution robustly (e.g., $1024px \times 768px $) at the fine structure, even trained around  $512px \times 512px $. (c) Existing SOTA methods could only work well on images in $1:1$ aspect ratio with the low-resolution format.}

    \label{imgs:teasing}
\end{center}}]

\newenvironment{alphafootnotes}
  {\par\edef\savedfootnotenumber{\number\value{footnote}}
  \renewcommand{\thefootnote}{\alph{footnote}}
  \setcounter{footnote}{0}}
  {\par\setcounter{footnote}{\savedfootnotenumber}}

\begin{alphafootnotes}
\let\thefootnote\relax\footnotetext{* Corresponding author}
\end{alphafootnotes}

\begin{abstract}
Face inpainting requires the model to have a precise global understanding of the facial position structure.
Benefiting from the powerful capabilities of deep learning backbones, recent works in face inpainting have achieved decent performance in ideal setting (square shape with $512px$). However, existing methods often produce a visually unpleasant result, especially in the position-sensitive details (e.g., eyes and nose), when directly applied to arbitrary-shaped images in real-world scenarios. The visually unpleasant position-sensitive details indicate the shortcomings of existing methods in terms of position information processing capability.
In this paper, we propose an \textbf{I}mplicit \textbf{N}eural \textbf{I}npainting \textbf{N}etwork (IN$^2$) to handle arbitrary-shape face images in real-world scenarios by explicit modeling for position information. Specifically, a downsample processing encoder is proposed to reduce information loss while obtaining the global semantic feature. A neighbor hybrid attention block is proposed with a hybrid attention mechanism to improve the facial understanding ability of the model without restricting the shape of the input. Finally, an implicit neural pyramid decoder is introduced to explicitly model position information and bridge the gap between low-resolution features and high-resolution output. Extensive experiments demonstrate the superiority of the proposed method in real-world face inpainting task.
\end{abstract}

\section{Introduction}
 Image inpainting aims to generate plausible content to complete the missing regions. Compared to image inpainting which focuses more on extracting relevant information from unmasked backgrounds, face inpainting requires the model to have a global understanding of the facial position structure, which is more challenging.
With the flourishing development of deep neural networks (DNNs)~\cite{krizhevsky2012imagenet} and generative adversarial networks (GANs)~\cite{goodfellow2014generative}, face inpainting has made significant progress. Specifically, previous works focus on the generated content and now can produce more realistic texture~\cite{xu2021texture} and diversified inpainting results~\cite{li2022mat} in ideal settings (e.g., $512px \times 512px $). However, the real-world settings from downstream applications are quite different from the ideal setting, where arbitrary-size inputs are what we have to cope with. Compared to the content and texture of general scenes, the position-sensitive face structure information is more affected by changes in image shape, which makes face inpainting in real-world scenarios more challenging.
When existing methods work in real-world scenarios, they can only accept specific inputs, as shown in Tab.~\ref{tab:input_limitation}. 

More importantly, when the shape and size change, they tend to produce the results with misaligned eyes and nose, as shown in Fig.~\ref{fig:summary} (b). It indicates existing methods are unable to form a satisfactory understanding of the position-sensitive facial structure in real-world scenarios, which results in a performance drop, as shown in Tab.~\ref{tab:performance_drop}.

Different inference settings and retraining have been tried to adapt previous methods to arbitrary-size input. By mitigating the gap in spatial structure information between training data and testing data, resizing can simply and effectively improve the performance of the model in most scenarios. Although extra resize processing is adopted, the duplicate artifacts are serious in the position-sensitive areas, such as eyes and nose, shown in Fig.~\ref{fig:summary} (d), which indicates the ineffective position information modeling~\cite{xu2021positional}.

 It is worth noticing that the arbitrary-shape image super-resolution (SR) task has been solved by representing an image with Implicit Neural Representation (INR). INR represents an object as a function that maps position to the corresponding signal value to model position information better~\cite{chen2021learning}. Compared with the SR task, face inpainting requires a larger receptive field for facial structure understanding to complete the masked areas. Therefore, how to leverage position information effectively to adapt the inpainting model to real-world face inpainting task is still an open problem. 

 In this paper, we develop a new INR inpainting model, capable of generating robust results for real-world face inpainting task due to its capacity to explicitly model position information. To deal with real-world high-resolution input, downsampling is an inevitable operation.
  Noticing the commonly used cascaded downsampling in the encoder may lead to information loss during the encoding process, we propose the \textbf{D}ownscale \textbf{P}rocessing \textbf{B}lock (DPB) to alleviate this problem. 
Although the popular window-based attention mechanism~\cite{vaswani2017attention} is beneficial for facial structure learning~\cite{li2022mat}, it restricts the acceptable inputs, which is not conducive to its application in real-world scenarios.
With neighborhood Attention~\cite{hassani2023neighborhood}, we propose a new variant of attention block to overcome this problem and learn better facial features, named \textbf{N}eighbor \textbf{H}ybrid \textbf{A}ttention \textbf{B}lock (NHAB). 
Finally, we propose an \textbf{I}mlpicit \textbf{N}eural \textbf{P}yramid \textbf{D}ecoder (INPD) to model the position information explicitly and mitigate the gap between low-resolution feature to high-resolution output. Thanks to the previous design, our model can directly deal with inputs of arbitrary size. Therefore, we design an\textbf{A}daptation \textbf{T}raining \textbf{S}trategy (ATS) which randomly crops input to irregular shapes to improve the facial spatial structure modeling ability of the model for the real-world challenge.
As shown in Fig.~\ref{imgs:teasing} (b), our method successfully restores arbitrary-shape input even in position-sensitive details. Our contributions are summarized as:
\begin{itemize}
    \item As far as we know, we are the first to propose the real-world face inpainting task from the perspective of spatial size and shape of input images. 
     
    \item Our proposed method IN$^2$ is a successful example of introducing INR into the field of face inpainting.
    
    \item Our method sets new state-of-the-arts on the benchmark dataset CelebA-HQ~\cite{karras2017progressive} in both ideal setting and real-world settings. 
\end{itemize}

\section{Related work}
\subsection{Image and Face Inpainting}
Image inpainting has been a long-standing problem in low-level computer vision, with mainstream methods falling into two categories. The first is traditional, non-learning-based image inpainting, which relies on strong low-level assumptions. For example, images are considered to have strong auto-correlation which allows the masked area to be filled with adjacent pixels~\cite{bertalmio2003simultaneous} or similar image patches~\cite{barnes2009patchmatch}. However, these methods tend to fail when the damaged area is relatively large.

The second category is learning-based methods. With the boom in deep learning techniques, convolutional neural networks~\cite{krizhevsky2012imagenet}, generative adversarial networks~\cite{goodfellow2014generative} and attention mechanism~\cite{vaswani2017attention} were successively proposed, which profoundly accelerates the development of inpainting. Many representative works~\cite{yu2018generative,zeng2019learning,suvorov2022resolution} have been proposed, which can produce decent restoration content in the ideal setting (e.g., $512px \times 512px$). When it comes to face inpainting, diversified results and controllable editing have been successfully addressed as recent hot topics~\cite{yildirim2023diverse}.
However, it is far from meeting the real-world needs that various aspect ratios and high-resolution outputs are required.
 

\subsection{Implicit Neural Representation}
Implicit Neural Representation (INR) is a novel image representation that models an object as a function defined in a continuous domain, mapping location to signal value. By explicitly modeling signals with position information, the feature can express beyond resolution limitations. Moreover, from a practical perspective, INR is a coordinate-based MLP, which is a memory-efficient framework for high-resolution data. With these advantages, INR has shown outstanding potential in many fields. 
For example, NeRF~\cite{martin2021nerf} employs implicit neural representation to perform novel view synthesis, which maps coordinates to RGB colors for a specific scene. LIIF and its subsequent methods ~\cite{chen2021learning,cao2023ciaosr,lte-jaewon-lee} propose a new solution for arbitrary super-resolution task by modeling an image as a function defined in a continuous domain. LINR ~\cite{linr} explores the feasibility of INR in low-level vision domains such as inpainting. CoordFill~\cite{liu2023coordfill} can directly convert input into an INR representation by their hyper-network for inpainting. However, existing INR inpainting methods can not achieve decent performance in real-world face inpainting settings. We believe the potential of INR has not been fully explored yet.

\section{Real-World Face Inpainting}
In this section, we explore the performance of existing methods in real-world scenarios. Firstly, we define the real-world scenarios and introduce our test setting. Secondly, we apply existing methods in real-world face inpainting task, which results in a huge performance drop. In the third part, we introduce our efforts to improve existing methods by extra processing and retraining. Finally, we summarize all experiments and introduce our motivation.


We here specifically confine real-world scenarios in two aspects: aspect ratio and resolution.
In terms of aspect ratio, we choose 3:4, 4:3 and 16:9 as representative ratios to conduct our experiments, for they are commonly used. As for resolution, under the condition of ensuring that the aspect ratio meets the requirements, we aim to make the resolution as close as possible to the original size of the images in the dataset to simulate the real-world demands in high-resolution scenarios. For example, to form our 16:9 (height: width) dataset, we crop the background area on both sides ($224px$ per side) of the image. It is worth emphasizing that the Lama in this paper was trained on $512px$ for fair evaluation purposes following~\cite{suvorov2022resolution}.

\subsection{Apply Existing Face Inpainting Methods to Real-World Scenarios}
Applying existing inpainting models to real-world scenarios, we find that existing methods have more or less limitations on the shape of the input image, as shown in Tab.~\ref{tab:input_limitation}. MAT~\cite{li2022mat} can only handle square input due to its window-based attention module, and CooordFill~\cite{liu2023coordfill} suffers the same limitation due to its patch-based hypernetwork architecture. Even the simplest conv-based methods require the resolution of the input to be an integer multiple of $2^{num}$, where num is the number of layers in the U-Net, due to the downsample and upsampline operations. We first consider the high-resolution inpainting task ($1024px\times 1024px$), where we can apply the existing model directly. As shown in Tab.~\ref{tab:performance_drop}, there is a huge performance drop between the ideal scenario ($512px\times512px$) and the high-resolution scenario with the origin inference setting.

\begin{table}[]
 \begin{center}
        \resizebox{\linewidth}{!}{
\begin{tabular}{c|c|c}
\hline
Methods   & Reasons for constraints    & Supported inputs             \\ \hline
Deepfill  & upsampling and downsampling in U-Net        & evenly divided by $2^{num}$  \\ \hline
LAMA    &  upsampling and downsampling in U-Net        & evenly divided by $2^{num}$ \\ \hline
MAT       & window-based attention  & square shape    \\ \hline
CoordFill & patch-based supernetwork & square shape        \\ \hline
\end{tabular}
}
\end{center}
\vspace{-0.10in}
\caption{\textbf{  Existing methods only accept specific shape inputs.}  }
    \label{tab:input_limitation}
    \vspace{-0.08in}
\end{table}

\begin{table}[]
 \begin{center}
        \resizebox{\linewidth}{!}{
\begin{tabular}{c|ccc|ccc|ccc}
\hline
Methods           & \multicolumn{3}{c|}{LaMa}                                                                             & \multicolumn{3}{c|}{MAT}                                                                              & \multicolumn{3}{c}{CoordFill}                                                                        \\ \hline
Setting     & \multicolumn{1}{c|}{$512\times512$} & \multicolumn{2}{c|}{$1024\times1024$} & \multicolumn{1}{c|}{$512\times512$} & \multicolumn{2}{c|}{$1024\times1024$} & \multicolumn{1}{c|}{$512\times512$} & \multicolumn{2}{c}{$1024\times1024$} \\ \hline
Inference         & \multicolumn{1}{c|}{origin}                      & \multicolumn{1}{c|}{origin}        & resize        & \multicolumn{1}{c|}{origin}                      & \multicolumn{1}{c|}{origin}        & resize        & \multicolumn{1}{c|}{origin}                      & \multicolumn{1}{c|}{origin}        & resize        \\ \hline
PSNR$\uparrow$    & \multicolumn{1}{c|}{29.079}                      & \multicolumn{1}{c|}{28.476}        & 29.014        & \multicolumn{1}{c|}{29.104}                      & \multicolumn{1}{c|}{27.739}        & 29.109        & \multicolumn{1}{c|}{29.336}                      & \multicolumn{1}{c|}{28.940}        & 28.962        \\ \hline
SSIM$\uparrow$    & \multicolumn{1}{c|}{0.913}                       & \multicolumn{1}{c|}{0.912}         & 0.915         & \multicolumn{1}{c|}{0.914}                       & \multicolumn{1}{c|}{0.908}         & 0.914         & \multicolumn{1}{c|}{0.893}                       & \multicolumn{1}{c|}{0.910}         & 0.911         \\ \hline
LPIPS$\downarrow$ & \multicolumn{1}{c|}{0.064}                       & \multicolumn{1}{c|}{0.089}        & 0.0816        & \multicolumn{1}{c|}{0.050}                       & \multicolumn{1}{c|}{0.073}        & 0.069        & \multicolumn{1}{c|}{0.077}                       & \multicolumn{1}{c|}{0.087}         & 0.086         \\ \hline
\end{tabular}
}
\end{center}
        \vspace{-0.10in}
\caption{\textbf{ Existing face inpainting models suffer from a performance drop in high-resolution scenarios and resizing can mitigate this problem.} Compared with the performance between ideal and high-resolution setting in direct inference setting, we can see that models deteriorate severely in high-resolution scenarios. Compared with different inference settings, we can see that extra resize processing can improve the performance.}
    \label{tab:performance_drop}
    \vspace{-0.08in}
\end{table}

\begin{table}[]
 \begin{center}
        \resizebox{\linewidth}{!}{
\begin{tabular}{c|cc|cc|cc}
\hline
Setting           & \multicolumn{2}{c|}{$768 \times 1024$ (3:4)} & \multicolumn{2}{c|}{$1024 \times 768$ (4:3)} & \multicolumn{2}{c}{$1024 \times 576$ (16:9)} \\ \hline
Inference         & \multicolumn{1}{c|}{origin}  & resize  & \multicolumn{1}{c|}{origin}  & resize  & \multicolumn{1}{c|}{origin}  & resize  \\ \hline
PSNR$\uparrow$    & \multicolumn{1}{c|}{29.073}  & \textbf{29.445}  & \multicolumn{1}{c|}{29.095}  & \textbf{29.449}  & \multicolumn{1}{c|}{29.942}  & \textbf{30.026}  \\ \hline
SSIM$\uparrow$    & \multicolumn{1}{c|}{0.916}   & \textbf{0.918}   & \multicolumn{1}{c|}{\textbf{0.916}}   & \textbf{0.916}   & \multicolumn{1}{c|}{\textbf{0.920}}   & 0.919   \\ \hline
LPIPS$\downarrow$ & \multicolumn{1}{c|}{0.085}   & \textbf{0.080}   & \multicolumn{1}{c|}{0.082}   & \textbf{0.080}   & \multicolumn{1}{c|}{\textbf{0.078}}  & 0.082   \\ \hline
\end{tabular}
}
\end{center}
        \vspace{-0.10in}
\caption{\textbf{ Inference with extra resize processing tends to be a superior inference setting in most scenarios.} We compared the performance of LaMa with and without additional processing in different aspect ratios and high-resolution settings. 'Resize' turns out to be a simple but effective way to improve the performance. }
    \label{tab:resize}
    \vspace{-0.15in}

\end{table}


\subsection{Adapt Existing Methods to Real-World Scenarios}

We speculate that the performance degradation is due to the distortion of existing methods in understanding the structure of facial position when the input size changes.
For the adaptation of existing face inpainting methods to real-world scenarios, we try to combine them with some extra resize processing. Specifically, we scale the input to the size of the training data and scale the output back to its origin size (high-resolution) to mitigate the gap in spatial structure information between training data and testing data. As shown in Tab.~\ref{tab:performance_drop}, with extra resize processing, all methods can achieve superior performance in high-resolution scenario.
we adopt extra resize processing in latter comparisons unless a specific explanation.

Intuitively, another method that can help existing methods maintain a decent understanding of facial structure and position in real-world scenarios is retraining them with arbitrary-shape input.
We use images with different aspect ratios whose total size is close to  $512px\times512px$ as training data to finetune LaMa, dubbed retrain, which is the same training strategy as our method (discussed in section 4). We test the origin model and the retrained model in different settings without extra processing. As shown in Tab.~\ref{tab:retrain}, training with different aspect ratio data can not improve the robustness of LaMa in a real-world setting. Moreover, it causes a performance drop in the ideal setting, which indicates that the existing model design cannot support learning robust facial representations in shape-changing training data.

\subsection{Summary and Motivation}
\begin{figure}[t]
    \begin{center}
        \includegraphics[width=1.0\linewidth]{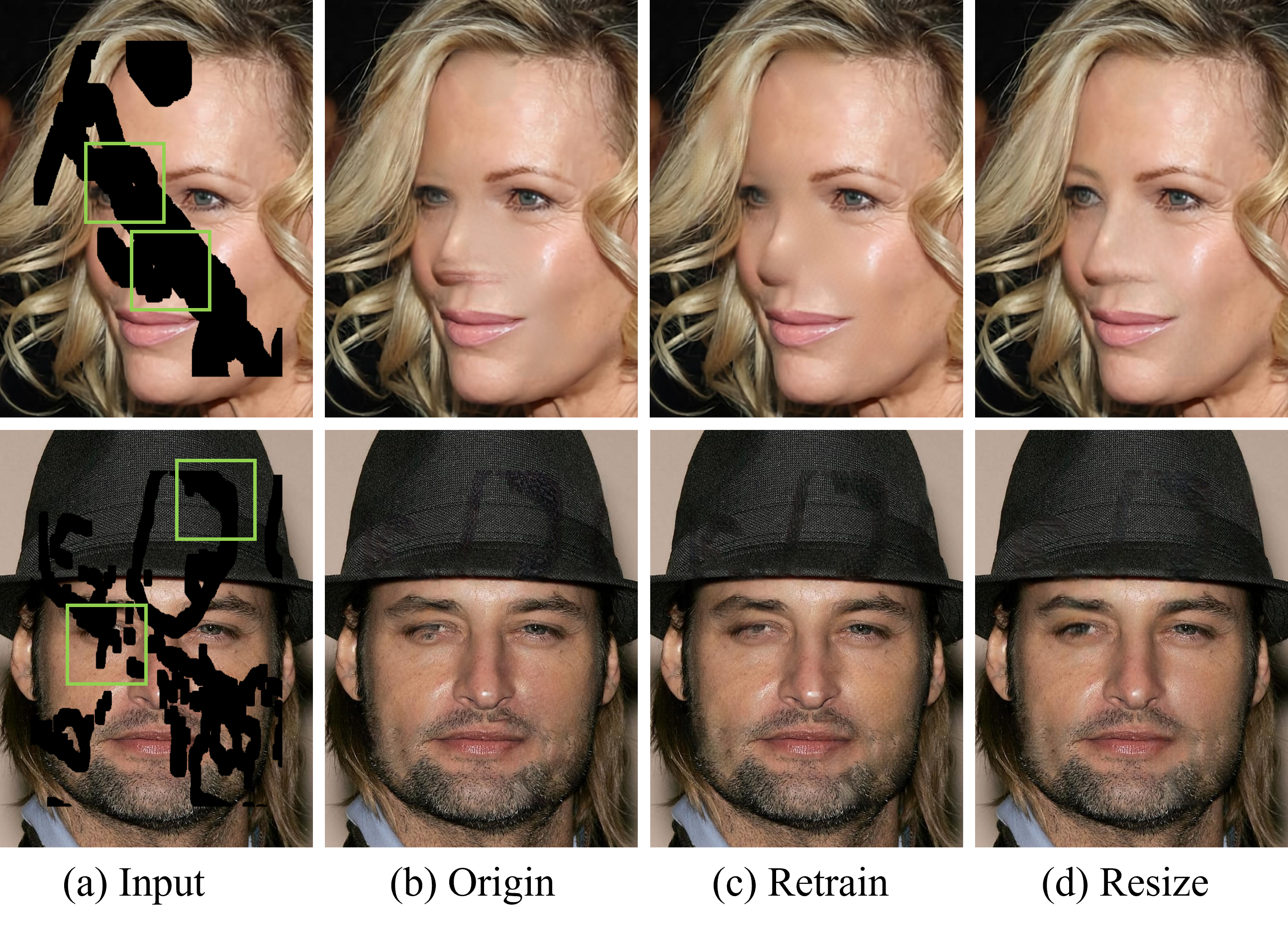}
    \end{center}
        \vspace{-0.13in}
    \caption{\textbf{Qualitative results of LaMa with various settings in a real-world setting ($1024px\times768px$).} Original LaMa suffers a performance drop in a real-world setting, which can be improved by extra resize processing. However, it is still unsatisfactory in position-sensitive details, such as eyes and mouth.} %
    \label{fig:summary}
        \vspace{-0.08in}
\end{figure}

So far, we now can output plausible content in real-world settings by extra resize processing in areas where position information is not sensitive (dubbed non-sensitive area), shown in Fig.~\ref{fig:summary} (d), which means we need to improve the capacity of modeling position for more realistic details.
Moreover, we notice training with different aspect ratio images can not improve performance in total but can mitigate the duplicate artifact, especially around the eyes and nose, as shown in Fig.~\ref{fig:summary} (b) and (c), which means training model with arbitrary-shape input is not a way without prospect.
Finally, as mentioned in \cite{xu2021positional}, duplicate artifact in conv-based networks is caused by implicit and weak modeling method for position information. With all the above observations, we propose a position-aware model with the same training strategy.



\begin{figure*}[t]
    \begin{center}
        \includegraphics[width=1.0\linewidth]{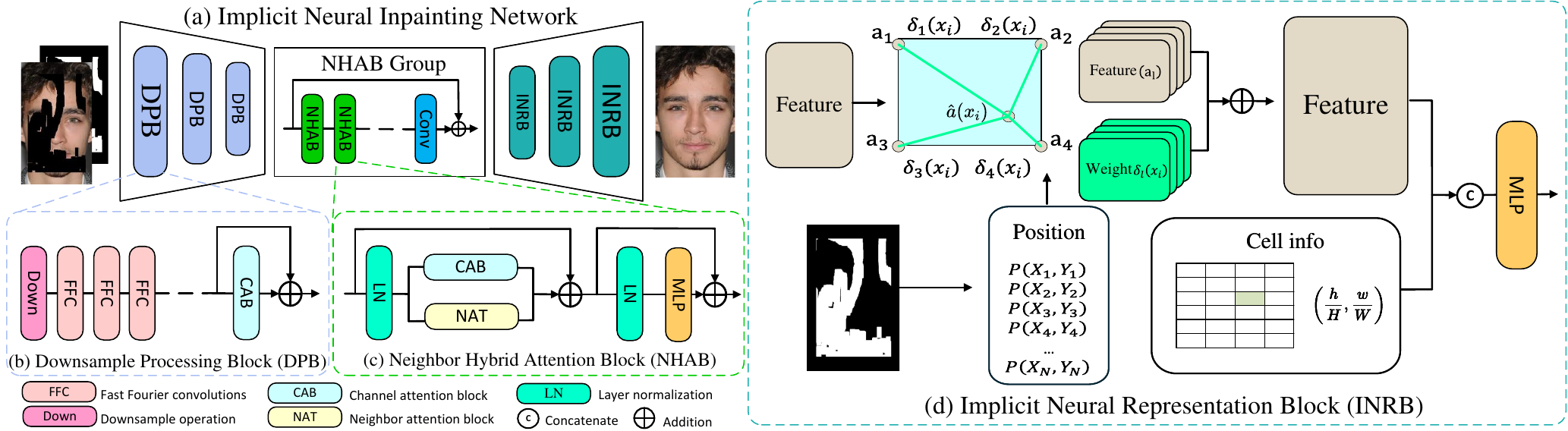}
    \end{center}
    \vspace{-0.10in}
    \caption{\textbf{The overview of the proposed implicit neural inpainting network for real-world face inpainting task.} (a) Implicit neural inpainting network consists of a downsample processing encoder, an attention body, and an implicit neural pyramid decoder. (b) We propose a downsample processing block to replace simple downsampling operation in other methods for more efficient encoding. (c) NHAB can overcome the limitation of window-based attention on image shape and enable superior facial structure learning. (d) Our implicit neural representation block can model position information explicitly for real-world task. 
    } %
    \label{fig:framework}
\vspace{-0.08in}
\end{figure*}

\section{Method}
In this section, we first introduce the overview of our proposed \textbf{I}mplicit \textbf{N}eural \textbf{I}npainting \textbf{N}etwork (IN$^2$), as shown in Fig.~\ref{fig:framework} (a). Then we illustrate the detailed design of each component. Finally, we discuss the training strategy and objective functions.

\begin{table}[]
 \begin{center}
        \resizebox{\linewidth}{!}{

\begin{tabular}{c|cc|cc|cc}
\hline
Setting           & \multicolumn{2}{c|}{$512\times512$} & \multicolumn{2}{c|}{$1024\times768$ (4:3)} & \multicolumn{2}{c}{$1024\times576$ (16:9)}     \\ \hline
training          & \multicolumn{1}{c|}{origin}           & retrain  & \multicolumn{1}{c|}{origin}            & retrain  & \multicolumn{1}{c|}{origin}          & retrain        \\ \hline
PSNR$\uparrow$    & \multicolumn{1}{c|}{\textbf{29.079}}  & 28.677   & \multicolumn{1}{c|}{\textbf{29.095}}   & 28.832   & \multicolumn{1}{c|}{\textbf{29.942}} & 29.819         \\ \hline
SSIM$\uparrow$    & \multicolumn{1}{c|}{\textbf{0.913}}   & 0.909    & \multicolumn{1}{c|}{\textbf{0.916}}    & 29.104   & \multicolumn{1}{c|}{\textbf{0.920}}  & \textbf{0.920} \\ \hline
LPIPS$\downarrow$ & \multicolumn{1}{c|}{\textbf{0.064}}   & 0.073    & \multicolumn{1}{c|}{\textbf{0.082}}    & 0.088    & \multicolumn{1}{c|}{\textbf{0.078}}  & 0.082          \\ \hline
\end{tabular}
}

\end{center}
        \vspace{-0.10in}
\caption{\textbf{ Although retrained with the arbitrary-shape dataset, the conv-based method still cannot adapt to real-world inpainting task.} We retrain LaMa with various aspect ratios data whose total number of pixels is equal to $512px \times 512px$, dubbed retrain. Compared with the origin model with resize processing, there is no obvious improvement. }
    \label{tab:retrain}
    \vspace{-0.10in}
\end{table}

\subsection{Overview}


To address real-world face inpainting task, our model is required to accept arbitrary image shapes and efficiently get a precise understanding of the facial position structure. Specifically, A large receptive field, a feature processing module that does not restrict input, and powerful positional information modeling capabilities are characteristics that we need to consider in model design. In response to the above issues, we have carefully designed the following modules.


 \begin{figure*}[t]
    \begin{center}
        \includegraphics[width=\linewidth]{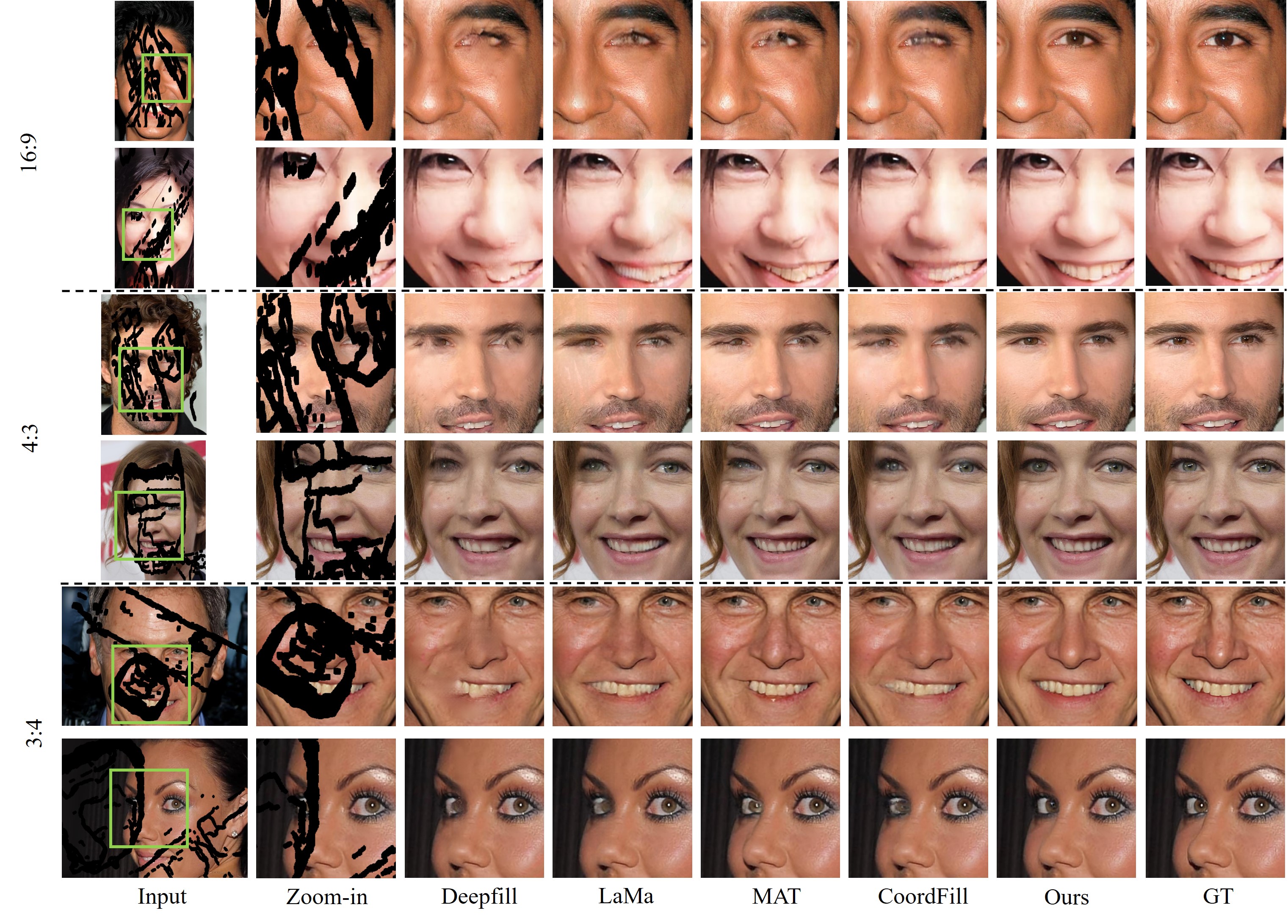}
    \end{center}
           \vspace{-0.15in}
    \caption{ \textbf{Qualitative comparison with different aspect ratios.} We display zoom-in results for easier comparison of position-sensitive details and whole results can be found in the appendix.  } %

    \label{fig:sota_celeba}
    \vspace{-0.20in}
\end{figure*}

\subsection{Implicit Neural Inpainting Network}

\subsubsection{Downsample Processing Encoder}
The face inpainting task inherently requires a precise understanding of the facial structure to produce masked content, demanding a large effective receptive field.
A cascade downsampling operation is a straightforward and popular solution~\cite{nazeri2019edgeconnect,zeng2022aggregated}. However, in the early encoding stage where the damaged area has not yet been repaired, it is unreasonable to conduct the same downsampling operation on both the damaged and intact areas. In practice, this may result in the information loss of intact areas and prevents the model from producing realistic detail for high-resolution applications. We propose a downsample processing block to alleviate this problem by providing additional feature processing in the early stage, which not only accelerates the restoration but also makes it easier for useful information in the features to be retained during the downsampling process.

Specifically, we choose Fast Fourier Convolutions (FFC) as our basic module which allows us to use global context in early layers by the power of fast Fourier transform~\cite{suvorov2022resolution}. As shown in Fig.~\ref{fig:framework} (b), we insert the FFC module behind the downsampling operation to retain more information. Moreover, we add a Channel Attention Block (CAB) to utilize global information.
 In practice, we stack three downsample processing blocks to form a downsample processing encoder that takes in the incomplete image concatenated with the given mask and produces features used for the later restoration.
 
\subsubsection{Neighbor Hybrid Attention Blocks}
More efficient feature processing modules can also bring better performance in face inpainting task~\cite{li2022mat}.
However, the simple self-attention method~\cite{vaswani2017attention} is not efficient and the window-based attention~\cite{liu2021swin} method contradicts the arbitrary image shape demanded in real-world scenarios. To address this problem, we adopt the neighborhood attention block (NAB)~\cite{hassani2023neighborhood} to avoid the restriction on resolution and process the feature efficiently.
Additionally, inspired by previous works that convolution can help Transformer get preferable visual representation and achieve easier optimization ~\cite{wu2021cvt,li2023uniformer}, we insert a Channel Attention Block (CAB) to aggregate global information, which is complementary to the local attention module. Following~\cite{chen2023hat}, we multiply the output of CAB with a small constant $\alpha$, to avoid the possible conflict of CAB and NAB on optimization and visual representation, the whole process is computed as 
\begin{gather} X_N={\rm LN}(X), \notag \\
  X_M={\rm \text{\rm NAB}}(X_N)+\alpha {\rm CAB} (X_N), \\
  Y={\rm MLP}({\rm LN}(X_M))+X_M, \notag 
\end{gather}
where $X_N$ and $X_M$ denote the intermediate features. $Y$ represents the output of the Neighbor Hybrid Attention Block (NHAB). 
We stack several NHAB to form an NHAB group for better capacity (as shown in Fig.~\ref{fig:framework} (c)).

 \begin{table*}[]
    \begin{center}
        \resizebox{\linewidth}{!}{


\begin{tabular}{c|ccc|ccc|ccc|ccc}
\hline
Settings       & \multicolumn{3}{c|}{$512 \times 512$}                                                          & \multicolumn{3}{c|}{$768 \times 1024$}                                                         & \multicolumn{3}{c|}{$1024 \times 768$}                                                         & \multicolumn{3}{c}{$1024 \times 576$}                                                         \\ \hline
Metrics        & \multicolumn{1}{c|}{PSNR$\uparrow$}  & \multicolumn{1}{c|}{SSIM$\uparrow$} & LPIPS$\downarrow$ & \multicolumn{1}{c|}{PSNR$\uparrow$}  & \multicolumn{1}{c|}{SSIM$\uparrow$} & LPIPS$\downarrow$ & \multicolumn{1}{c|}{PSNR$\uparrow$}  & \multicolumn{1}{c|}{SSIM$\uparrow$} & LPIPS$\downarrow$ & \multicolumn{1}{c|}{PSNR$\uparrow$}  & \multicolumn{1}{c|}{SSIM$\uparrow$} & LPIPS$\downarrow$ \\ \hline
Deepfill       & \multicolumn{1}{c|}{26.677}          & \multicolumn{1}{c|}{0.898}          & 0.109             & \multicolumn{1}{c|}{28.507}          & \multicolumn{1}{c|}{0.917}          & 0.097             & \multicolumn{1}{c|}{29.742}          & \multicolumn{1}{c|}{0.916}          & 0.083             & \multicolumn{1}{c|}{29.084}          & \multicolumn{1}{c|}{0.918}          & 0.100             \\ \hline
LAMA           & \multicolumn{1}{c|}{29.079}          & \multicolumn{1}{c|}{0.913}          & 0.064             & \multicolumn{1}{c|}{29.445}          & \multicolumn{1}{c|}{0.918}          & 0.080             & \multicolumn{1}{c|}{29.448}          & \multicolumn{1}{c|}{0.915}          & 0.080             & \multicolumn{1}{c|}{30.026}          & \multicolumn{1}{c|}{0.918}          & 0.082             \\ \hline
MAT            & \multicolumn{1}{c|}{29.104}          & \multicolumn{1}{c|}{0.914}          & 0.050             & \multicolumn{1}{c|}{29.392}          & \multicolumn{1}{c|}{0.916}          & 0.068             & \multicolumn{1}{c|}{30.001}          & \multicolumn{1}{c|}{\textbf{0.923}} & 0.066             & \multicolumn{1}{c|}{29.743}          & \multicolumn{1}{c|}{0.917}          & 0.069             \\ \hline
CoordFill      & \multicolumn{1}{c|}{29.336}          & \multicolumn{1}{c|}{0.893}          & 0.077             & \multicolumn{1}{c|}{29.801}          & \multicolumn{1}{c|}{0.919}          & 0.083             & \multicolumn{1}{c|}{29.209}          & \multicolumn{1}{c|}{0.913}          & 0.067             & \multicolumn{1}{c|}{29.743}          & \multicolumn{1}{c|}{0.917}          & 0.086             \\ \hline
IN$^{2}$(Ours) & \multicolumn{1}{c|}{\textbf{30.285}} & \multicolumn{1}{c|}{\textbf{0.927}} & \textbf{0.044}    & \multicolumn{1}{c|}{\textbf{30.391}} & \multicolumn{1}{c|}{\textbf{0.924}} & \textbf{0.056}    & \multicolumn{1}{c|}{\textbf{30.040}} & \multicolumn{1}{c|}{0.919}          & \textbf{0.061}    & \multicolumn{1}{c|}{\textbf{31.597}} & \multicolumn{1}{c|}{\textbf{0.929}} & \textbf{0.051}    \\ \hline
\end{tabular}
}
    \end{center}
        \vspace{-0.10in}
    \caption{\textbf{ Quantitative results in different settings.} We achieve superior performance in both ideal settings and real-world settings. }
    \label{tab:sota_celeba}
    \vspace{-0.12in}
\end{table*}

\subsubsection{Implicit Neural Pyramid Decoder}
Position information is crucial for handling input images of different shapes in the real-world scenarios~\cite{chen2021learning}. We propose an implicit neural representation block to explicitly model position information and upsample feature compressed for a large receptive field, as shown in Fig.~\ref{fig:framework} (d). With the above module, we can establish a relationship between positional coordinates and features. Specifically, to predict the feature $\hat{a}\left ( x_{i}  \right )$ at an arbitrary query coordinate $ x_{i}$, We gather its neighborhood information. The entire process is computed as
\begin{align} 
    \hat{a}\left ( x_{i}  \right )  =  {\textstyle \sum_{l}^{}  f \left ( a_{l} , \delta _{l}\left ( x_{i}  \right )   \right ) },
\end{align}
where $l$ is the point index in the local region centered at the query coordinate $x_{i}$, ( $l$ can be top-left, top-right, bottom-left and bottom-right), and $ \delta _{l}\left ( x_{i}  \right )$ is the weight of the neighboring feature value $a_{l}$. $ \delta _{l}\left ( x_{i}  \right )$ is the distance between $x_{i}$ and the position of $a_{l}$. The feature is upsampled during the above process.
Moreover, to adapt to various image shapes better, we introduce the cell information which specifies the height and width of the query pixel~\cite{chen2021learning}, which is defined as
\begin{align}
\left ( c_{h},c_{h} \right ) & = \left ( \frac{h}{H},\frac{w}{W}  \right ) ,
\end{align}
where h, w represents the height and width of the feature, and H, W represents the height and width of output respectively.

However, the gap between the high-resolution output demanded by real-world downstream applications and the low-resolution feature required for a large receptive field is not negligible. One-step upsampling is not an appropriate solution, which leads to a decrease in model performance~\cite{cdpn}.
In light of this, we propose an implicit neural pyramid decoder to produce filled images progressively.
Specifically, we set up a three-layer pyramid, where each layer is an INRB module that operates on its resolution and only the final layer operates on the target one. 
The whole process is computed as 
\begin{align} 
a_{ \textup{layer}+1}\left ( x_{i}  \right )  = MLP\left ( \hat{a}_{ \textup{layer}}\left ( x_{i} \right ) ,\left ( c_{h},c_{w} \right ) \right ),
\end{align}
where $a_{ \textup{layer}+1}\left ( x_{i}  \right )$ is feature value at the coordinate $x_{i}$ in the pyramid, layer indicates which layer of the pyramid the feature belongs to and $\left ( c_{h}, c_{w}  \right )$ are the cell information, which specifies the height and width of the query pixel.

\subsection{Training Details}

\subsubsection{Adaptive Training Strategy}
We want to train models on data with different aspect ratios to achieve robust facial structure modeling capabilities. Specifically, we randomly crop the marginal area to create datasets with different aspect ratios while maintaining the total pixels of each image is close to $512px \times 512px$. Compared to the result in Tab.~\ref{tab:retrain}, Our model design enables it to learn robust facial structure modeling ability on data with drastic changes in spatial positional information. The ablation study validates that our strategy is simple and effective.

\subsubsection{Training Objective}
 We follow CoordFill~\cite{liu2023coordfill} to train our model with different perception losses, including saturating adversarial loss~\cite{goodfellow2014generative}, perceptual loss~\cite{johnson2016perceptual} and feature matching loss~\cite{wang2018high}. Besides, we also use the $R_{1}$ regularization~\cite{mescheder2018training}, written as $R_{1}=E_{x} \left \| \nabla D(x)\right \|$, to stabilize the training. 
We calculate the adversarial loss as
	\begin{align}
		\mathcal{L}_{\rm G} & = - \mathbb{E}_{\hat x} \left[ \log \left(D \left( \hat x \right) \right) \right] \,, \\
		\mathcal{L}_{\rm D} & = - \mathbb{E}_{x} \left[ \log \left( D \left( x \right) \right) \right] - \mathbb{E}_{\hat x} \left[ \log \left( 1 - D \left( \hat x \right) \right) \right] \,,
	\end{align}
	where $x$ and $\hat x$ are the real and generated images. The perceptual loss is formulated as
	\begin{align}
		\mathcal{L}_{per} = \sum_{i}{ \tau \left\| \phi_{i} \left( \hat x \right) - \phi_{i} \left ( x \right) \right\|_{1} } \,,
	\end{align}
	where $\phi_{i}(\cdot)$ denotes the layer activation of a pre-trained VGG-19~\cite{simonyan2014very} network. $\tau$ calculates the differences in the feature domain via $L_1$ loss. 
     Next, the feature matching loss is adopted for stabilizing the GAN training~\cite{liu2023coordfill,suvorov2022resolution} and can be formulated as
    \begin{equation}
    L_{fm} = \sum_{i} \tau \left(D^{i}\left(x\right), D^{i}\left(\hat x \right)\right),
    \end{equation} where $D^{i}$ denotes the activations from the $i$-th layer of the discriminator $D$. The overall loss of the generator is
	\begin{align}
		L_{total} =  L_{G}  +  \lambda_{per} L_{per} + \lambda_{fm} L_{fm} + \lambda_{R} R_{1} .
	\end{align}
	where $\lambda_{per}=10$, $\lambda_{fm}=0.1$ and $ \lambda_{R}= 10$ .

\section{Experiments}
In this section, implementation details, dataset and evaluation metrics are first introduced. Then we conduct qualitative and quantitative comparative experiments with SOTA methods in both ideal and real-world scenarios. Additionally, we compare our method with existing methods with super-resolution networks as post-processing, as it is an intuitive solution that can handle real-world scenarios. Moreover, we introduce an ablation study. Due to limited layout, latency and model size information can be found in the appendix.

\subsection{Implementation Details}
All experiments are carried out on 2 NVIDIA RTX 4090 GPUs. All models are trained for 100 epochs with batch size 3, and the learning rate decays by cosine annealing after a warm-up phase of 15 epochs. We use the Adam~\cite{kingma2014adam} optimizer with an initial learning rate $4 \times 10^{-5}$ and the maximum $4 \times 10^{-4}$.

In our framework, the number of the pyramid layer is three and the convolution channel dimensions for the encoder and decoder are 256, 128, and 64, respectively. The attention body contains two neighbor hybrid attention groups. For each of them, there are four neighbor hybrid attention blocks. Empirically, we set the $\alpha$ as a small constant number (0.03) following \cite{chen2023hat}.


\subsection{Datasets and Metrics}
We experiment on the CelebA-HQ~\cite{karras2017progressive} dataset. We use the free-form masks~\cite{liu2018image} for training and testing, while the masked area is $ 20 \% - 30 \%$ following common settings~\cite{suvorov2022resolution,liu2023coordfill}.

As for metrics, we follow previous works~\cite{suvorov2022resolution,liu2023coordfill} to use PSNR, SSIM~\cite{wang2004image} and LPIPS~\cite{zhang2018unreasonable}, because they can evaluate images from different perspectives.

\subsection{Comparison with SOTAs}
	We compare the performance of our method and other SOTA methods under different aspect ratios of input in both high-resolution and low-resolution scenarios. For a fair comparison, we use publicly available model weights to test on the same masks, except for LaMa. Because the public weight is trained on $256px \times 256px$, we train it on $512px \times 512px$ for comparison. 
 It is worth emphasizing that all comparison methods, except for LaMa, adopt extra processing to meet the requirements of the model and enhance performance under non-1:1 aspect ratio.  When the resolution of the test set is close to $512px$, LaMa do not need extra processing. Due to layout limitations, we only present the results of ideal scenario and high-resolution scenarios with different aspect ratios here. More results (e.g., low-resolution with different aspect ratios) can be found in the appendix. 

 \subsubsection{Quantitative Comparisons}
For the settings in previous work ~\cite{suvorov2022resolution} (dubbed ideal setting), we conduct the experiments at $512px \times 512px$.
As for real-world settings, we conduct experiments in three representative aspect ratios with high-resolution format, including $768px \times 1024px$, $1024px \times 768px$, and $1024px \times 576px$. 
  As illustrated in Tab.~\ref{tab:sota_celeba}, no matter the ideal setting and real-world settings, our method achieves state-of-the-art performance. More experimental results are provided in the appendix.

    \subsubsection{Qualitative Comparisons}
    We show the qualitative inpainting results in real-world scenarios (different aspect ratios in high-resolution format) in Fig.~\ref{fig:sota_celeba}. To focus on the position-sensitive details, where existing methods fail, such as eyes and mouth, we post the zoom-in results. The whole image results are provided in the appendix. Compared with other methods, our method achieves superior performance in all settings.

\begin{table}[]
 \begin{center}
        \resizebox{\linewidth}{!}{
\begin{tabular}{c|cc|cc}
\hline
Settings & \multicolumn{2}{c|}{(a)Real-world~($1024\times576$)}    & \multicolumn{2}{c}{(b)High-resolution~($1024\times1024$)}       \\ \hline
Model    & \multicolumn{1}{c|}{MAT+LIIF} & IN$^{2}$    & \multicolumn{1}{c|}{MAT+GFPGAN} & IN$^{2}$    \\ \hline
PSNR$\uparrow$     & \multicolumn{1}{c|}{29.531}    & \textbf{31.496} & \multicolumn{1}{c|}{28.758}     & \textbf{29.683} \\ \hline
SSIM$\uparrow$     & \multicolumn{1}{c|}{0.914}    & \textbf{0.936}  & \multicolumn{1}{c|}{0.910}      & \textbf{0.929}  \\ \hline
LPIPS$\downarrow$    & \multicolumn{1}{c|}{0.070}    & \textbf{0.052}  & \multicolumn{1}{c|}{0.085}      & \textbf{0.062}  \\ \hline
\end{tabular}
}
\end{center}
        \vspace{-0.10in}
\caption{\textbf{Comparisons with SR-based Post-Processing.} We compare our IN$^{2}$ with SOTA inpainting model plus the super-resolution model in real-world and high-resolution settings respectively. The results indicate our method has better performance.}
    \label{tab:sr_post}
        \vspace{-0.05in}
\end{table}

\subsection{Comparisons with SR-based Post-Processing}
A simple and straight alternative to our method is to resize the input to comply with the training data and use a super-resolution model to scale the inpainting output back to its original size. We conduct this comparison in the real-world ($1024px\times576px$) and high-resolution settings ($1024px\times 1024px$) respectively.
For a fair comparison, we choose MAT~\cite{li2022mat} (CVPR 22 best paper finalist) as the inpainting backbone for its excellent performance. We choose LIIF~\cite{chen2021learning} (arbitrary SR model) for real-world setting and GFPGAN~\cite{wang2021towards} (human face SR model) for high-resolution setting. As shown in Tab.\ref{tab:sr_post} (a) and (b), our method has better performance because we internalize the position modeling ability of INR into the inpainting network. Although the extra SR network can improve the resolution, it cannot preserve the original image texture, which results in bad evaluation scores.

\subsection{Ablation Study}
	\label{sec:abl}
	In this section, we tease apart components in our framework and evaluate them in a real-world scenario ($1024px \times 576px$). The quantitative comparison is shown in Tab.~\ref{tab:ablation}.
 Specifically, we replace the adaptive training strategy with normal training data ($512px\times 512px$). We use the encoder of LAMA\cite{suvorov2022resolution} to replace DPB and increase its parameters for fair comparison and set the layer
number of the pyramid to 1 to validate the effectiveness of our pyramid. As for the attention module, we substitute it with more FFC modules, which is a common choice in networks designed without attention mechanisms \cite{suvorov2022resolution,jain2023keys}.
 The results demonstrate that our proposed modules can improve the receptive field, semantic learning and enhance the model’s generalization respectively, which is further beneficial for the overall performance.
 
According to the results of the ablation experiment, part of the improvements in our model come from the training strategy. It is worth emphasizing that without our INR structure, the model cannot learn robust inpainting capacity under the same training strategy, as shown in Tab.~\ref{tab:retrain}.

\begin{table}[]
 \begin{center}
        \resizebox{\linewidth}{!}{
\begin{tabular}{c|c|c|c|c|c|c}
\hline
Adaptive
Training Strategy & DBP           & NHAB          & Pyramid       & PSNR$\uparrow$ & SSIM $\uparrow$ & LPIPS $\downarrow$ \\ \hline
              &               &               &               & 29.690         & 0.916           & 0.074              \\ \hline
$\checkmark$  &               &               &               & 31.174         & 0.926           & 0.063              \\ \hline
$\checkmark $ & $\checkmark $ &               &               & 31.321         & 0.928           & 0.059              \\ \hline
$\checkmark $ & $\checkmark $ & $\checkmark $ &               & 31.363         & 0.935           & 0.055              \\ \hline
$\checkmark $ & $\checkmark $ & $\checkmark $ & $\checkmark $ & 31.496         & 0.936           & 0.052              \\ \hline
\end{tabular}
}
\end{center}
        \vspace{-0.10in}
\caption{\textbf{ Ablation experiments.} As we insert the module we design, the performance of the model is constantly improving, which indicates the effectiveness of our proposed module. }
    \label{tab:ablation}
        \vspace{-0.12in}
\end{table}




\section{Conclusion}
In this paper, we focus on real-world face inpainting task (e.g., high-resolution and arbitrary aspect ratio) and find that existing methods have significant shortcomings in terms of both acceptable inputs and performance in real-world scenarios. Targeting the task characteristics that require a precise understanding of the facial position structure, we try to adapt existing methods to real-world scenarios with extra resize processing and retraining. However these methods still fail to achieve satisfactory results due to their limited ability to model position information. To address this issue, we propose an implicit neural network (IN$^{2}$). With our well-designed modules, it can explicitly model facial position structure and accept arbitrary-size input in real-world face inpainting scenarios. Our experiments show the robust performance of the proposed method in several real-world settings compared with several state-of-the-art methods.

\appendix

\section{Overview}
In this appendix, we supplement a comparison of extra processing methods and a comparative experiment in different real-world settings. Moreover, we add the whole image corresponding to the zoom-in results in the main text. Finally, we list the latency and model size comparison.

\section{Comparison of Extra Processing Methods}
For SOTA methods that require square input (e.g., MAT), we test another extra processing method. We scale the long edge of the input to $512px$ and then perform padding operations, dubbed pad. Moreover, we try different pad methods, including constant pad and edge pad. We compare pad and the resize processing in the main text in real-world setting with a relative low-resolution format ($683px \times 384px$), as shown in Tab.~\ref{tab:pad}. Extra resize processsing tends to be better choice, and we adopt this method for fair comparison.

\begin{table}[h]
\renewcommand\arraystretch{1.1}
 \begin{center}
        \resizebox{\linewidth}{!}{

\begin{tabular}{c|c|c|c}
\hline
Inference Settings & pad-constant & pad-edge & resize          \\ \hline
PSNR$\uparrow$              & 29.448    & 29.715   & \textbf{29.991} \\ \hline
SSIM $\uparrow$             & 0.908     & 0.910    & \textbf{0.919}  \\ \hline
LPIPS$\downarrow$             & 0.067    & 0.065    & \textbf{0.060}  \\ \hline
\end{tabular}
}
 \end{center}
      \vspace{-0.10in}
 \caption{\textbf{ Comparison of extra processing methods.} Extra resize processing tends to be better. }
     \vspace{-0.15in}
     \label{tab:pad}
\end{table}

\section{Quantitative Comparisons in More Settings}
We conduct comparative experiments between our method and the main state-of-the-art methods under multiple settings, including the pure high-resolution setting ($1024px \times 1024px$) and low-resolution settings with various aspect ratios. As for the low-resolution settings, we keep the total number of pixels in the image equal to $512 \times 512$ while keeping the aspect ratio of the input constantly changing, which helps us see the impact on model performance when the aspect ratio changes. As shown in Table.~\ref{tab:more}, our method achieves better performance in all settings.


 \begin{figure*}[t]
    \begin{center}
        \includegraphics[width=1.0\linewidth]{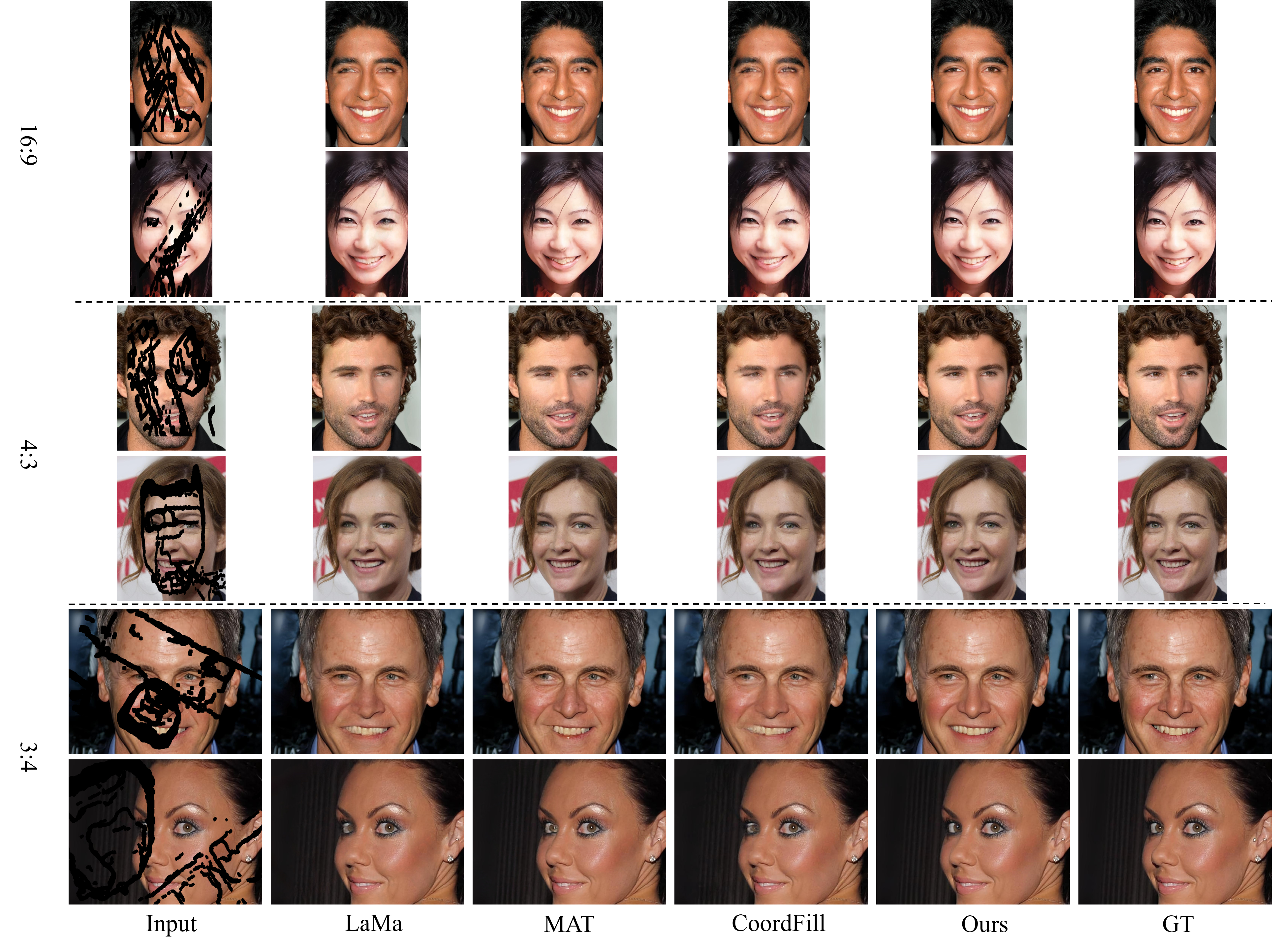}
    \end{center}
        \vspace{-0.05in}
    \caption{\textbf{Qualitative comparison with different aspect ratios.}} %
    \label{fig:sup}
        \vspace{-0.08in}
\end{figure*}


\section{More Visual Results}
To better demonstrate the inpainting performance in details where are sensitive to the change of position information, such as eyes and nose, we only post the zoom-in results in the main paper. We supplement some whole results in this section. To make it easier to compare the differences between different methods, we only list part of methods that achieve better performance.

\begin{table}[h]
\renewcommand\arraystretch{1.1}
 \begin{center}
        \resizebox{\linewidth}{!}{
\begin{tabular}{c|c|c|c|c}
\hline
  Methods    & LaMa & MAT   & CoordFill& $IN^{2}$(Ours)  \\ \hline
Param & 27.3 M   & 61.6M & 31.6M     & 15.5M \\ \hline
Latency & 13ms    & 28ms  & 10ms      & 20ms  \\ \hline
\end{tabular}

}
 \end{center}
      \vspace{-0.10in}
 \caption{\textbf{ Model size and the inference time.} We test inference time  at  $512px \times 512px$ in RTX 4090 }
     \vspace{-0.15in}
     \label{tab:model}
\end{table}

\section{Model Size and Latency}
We test the inference time of mentioned SOTA methods and our method at $512px \times 512px$ in RTX 4090 and add the model information, as shown in Table 2. At similar inference speeds, our model has fewer parameters. Due to the bottleneck of our computing speed being attention operations, our inference time can be shorter in the future.

\begin{table*}[]
 \begin{center}
        \resizebox{\linewidth}{!}{
\begin{tabular}{c|ccc|ccc|ccc|ccc}
\hline
\multirow{2}{*}{Methods} & \multicolumn{3}{c|}{$1024 \times 1024$}                                                        & \multicolumn{3}{c|}{$443 \times 591$}                                                          & \multicolumn{3}{c|}{$591 \times 443$}                                                          & \multicolumn{3}{c}{$683 \times 384$}                                                          \\ \cline{2-13} 
                         & \multicolumn{1}{c|}{PSNR$\uparrow$}  & \multicolumn{1}{c|}{SSIM$\uparrow$} & LPIPS$\downarrow$ & \multicolumn{1}{c|}{PSNR$\uparrow$}            & \multicolumn{1}{c|}{SSIM$\uparrow$}           & LPIPS$\downarrow$ & \multicolumn{1}{c|}{PSNR$\uparrow$}            & \multicolumn{1}{c|}{SSIM$\uparrow$}           & LPIPS$\downarrow$ & \multicolumn{1}{c|}{PSNR$\uparrow$}            & \multicolumn{1}{c|}{SSIM$\uparrow$}           & LPIPS$\downarrow$ \\ \hline
LAMA                     & \multicolumn{1}{c|}{29.014}          & \multicolumn{1}{c|}{0.915}          & 0.082             & \multicolumn{1}{c|}{29.516}          & \multicolumn{1}{c|}{0.915}          & 0.065             & \multicolumn{1}{c|}{29.463}          & \multicolumn{1}{c|}{0.913}          & 0.064             & \multicolumn{1}{c|}{30.330}          & \multicolumn{1}{c|}{0.920}          & 0.063             \\ \hline
MAT                      & \multicolumn{1}{c|}{29.109}          & \multicolumn{1}{c|}{0.914}          & 0.069             & \multicolumn{1}{c|}{29.634}          & \multicolumn{1}{c|}{0.918}          & 0.056             & \multicolumn{1}{c|}{29.454}          & \multicolumn{1}{c|}{0.916}          & 0.056             & \multicolumn{1}{c|}{29.991}          & \multicolumn{1}{c|}{0.919}          & 0.060             \\ \hline
CoordFill                & \multicolumn{1}{c|}{28.962}          & \multicolumn{1}{c|}{0.911}          & 0.087             & \multicolumn{1}{c|}{29.624}          & \multicolumn{1}{c|}{0.916}          & 0.074             & \multicolumn{1}{c|}{29.579}          & \multicolumn{1}{c|}{0.914}          & 0.074             & \multicolumn{1}{c|}{29.978}          & \multicolumn{1}{c|}{0.917}          & 0.082             \\ \hline
$IN^{2}$(Ours)           & \multicolumn{1}{c|}{\textbf{29.683}} & \multicolumn{1}{c|}{\textbf{0.929}} & \textbf{0.062}    & \multicolumn{1}{c|}{\textbf{30.508}} & \multicolumn{1}{c|}{\textbf{0.928}} & \textbf{0.045}    & \multicolumn{1}{c|}{\textbf{30.124}} & \multicolumn{1}{c|}{\textbf{0.921}} & \textbf{0.047}    & \multicolumn{1}{c|}{\textbf{31.864}} & \multicolumn{1}{c|}{\textbf{0.935}} & \textbf{0.051}    \\ \hline
\end{tabular}
}
\end{center}
        \vspace{-0.08in}
\caption{\textbf{ Quantitative results in different settings.} We achieve superior performance in all settings. }
    \label{tab:more}
        \vspace{-0.08in}
\end{table*}

\bibliographystyle{named}
\bibliography{ijcai24}

\end{document}